\documentclass{article}
\usepackage{spconf,amsmath,graphicx}
\usepackage{float}
\usepackage[dvipsnames]{xcolor}
\usepackage{soul}
\usepackage{tabularx}
\usepackage{multirow}
\usepackage{placeins}
\usepackage{pgfplots}
\pgfplotsset{compat=newest}
\usepackage{subcaption}
\usepackage{stfloats}
\usepackage{hyperref}
\usepackage{todonotes}

\title{TRScore: A Novel GPT-based Readability Scorer for ASR Segmentation and Punctuation model evaluation and selection}

\name{
\begin{tabular}{@{}c@{}}
Piyush Behre*\qquad 
Sharman Tan*\qquad 
Amy Shah\qquad
Harini Kesavamoorthy\\
Shuangyu Chang\qquad 
Fei Zuo\qquad
Chris Basoglu\qquad
Sayan Pathak\sthanks{These authors have contributed equally.}
\end{tabular}
}
\address{Microsoft Corporation}

\begin{document}

\maketitle

\begin{abstract}
Punctuation and Segmentation are key to readability in Automatic Speech Recognition (ASR), often evaluated using F$_1$ scores that require high-quality human transcripts and do not reflect readability well. Human evaluation is expensive, time-consuming, and suffers from large inter-observer variability, especially in conversational speech devoid of strict grammatical structures. Large pre-trained models capture a notion of grammatical structure.  We present TRScore, a novel readability measure using the GPT model to evaluate different segmentation and punctuation systems. We validate our approach with human experts. Additionally, our approach enables quantitative assessment of text post-processing techniques such as capitalization, inverse text normalization (ITN), and disfluency on overall readability, which traditional word error rate (WER) and slot error rate (SER) metrics fail to capture. 
TRScore is strongly correlated to traditional F$_1$ and human readability scores, with Pearson's correlation coefficients of 0.67 and 0.98, respectively. It also eliminates the need for human transcriptions for model selection.

\end{abstract}
\section{Introduction}
\label{sec:intro}
Automatic Speech Recognition (ASR) is gaining prominence with many different applications ranging from command-and-control scenarios (e.g., Alexa, Hey Siri, Ok Google, Cortana) to continuous speech recognition (e.g., live / closed captions, dictation, call center transcriptions, etc.). Word Error Rate (WER) and latency metrics dominate the field despite their limited focus on overall readability of ASR outputs.
 
A vast majority of ASR test sets lack written form transcriptions, limiting research on readability. 
Punctuation metrics like F$_1$ must rely on human transcriptions for ground truth labels. Any human involvement is expensive, time-consuming, non-scalable, and prone to high inter-observer variability, especially as text deviates from read-out speech to more spontaneous speech.
Furthermore, traditional metrics do not necessarily capture readability; systems with different precision and recall values may share similar F$_1$ scores.

Automated evaluations thus far generally fall into the categories of supervised regression, supervised ranking, and unsupervised matching. Supervised regression and ranking methods such as BLEURT \cite{sellam2020bleurt}, COMET \cite{rei2020comet}, and BEER \cite{stanojevic2014beer} train models to predict human judgments. However, such approaches require human judgments for supervised learning.

Unsupervised matching frameworks include BLEU \cite{papineni2002bleu} and chrF \cite{popovic-2015-chrf} (for machine translation quality evaluation), ROUGE \cite{rouge} (for semantic coverage evaluation of summarization), and metrics such as MoverScore \cite{zhao2019moverscore}, BERTScore \cite{zhang2019bertscore}, and BARTScore \cite{yuan2021bartscore} for evaluation of text generation using contextual embeddings.
While these metrics are powerful, they still require reference text.
SemDist \cite{kim2021evaluating} uses semantic distance to measure the semantic correctness of ASR outputs, noting the limitations of WER.
SemDist focuses on the impact of recognition errors on semantics and has been shown to have high correlation with human evaluation for NLU tasks.

An oracle (intelligent machine) that scores like a human expert and reports a single number measuring readability of a passage would yield several advantages:
\begin{itemize}
   \item {\bf Time advantage}: Measure the readability of ASR output in few seconds (enabling rapid experimentation),
   \item {\bf Consistency advantage}: Get more consistent results between runs with low variability in the scores, and 
   \item {\bf Cost advantage}: Significantly lower the cost of scoring compared to human experts.
\end{itemize}

In this paper, we present Transcription Readability Score (TRScore) as a close proxy of such an oracle.
To our knowledge, we are the first to capture the importance of display elements such as segmentation and punctuation, while enabling fast and accessible readability scoring with no reference text.

Large pre-trained language models (L-PLMs) capture vast knowledge from written text on the web. They are often used for text generation, text summarization, and question \& answering tasks. But L-PLMs are also capable of interpreting word sequences and generating likelihood scores. Further, they can be fine-tuned to understand custom or enterprise domain terminology. TRScore leverages L-PLMs to generate readability score distributions on corpora.
We make our TRScore tool available through Azure Cognitive Services Open AI service \cite{cognitiveopenai}. While we currently leverage GPT models \cite{brown2020language}, our methodology also works with other L-PLMs.

\section{Methods}
\subsection{Transcription Readability Score (TRScore)}

We leverage L-PLMs as the base model and use it in zero-shot mode for our scorer. We hypothesize that the L-PLMs generate a lower negative likelihood score for a well formed linguistically/grammatically correct sentence. As a sentence deviates from its correct form the negative likelihood score increases indicating deteriorating readability.   
We aggregate the negative log likelihood probabilities for all the words (including punctuation) in a sentence and derive the absolute score for that sequence. A percentile distribution of scores across the entire corpus represents the aggregate scoring.

Readability is presented as a relative improvement of this score with respect to a baseline (or reference text) instead of an absolute score. For instance, using GPT3 model\cite{brown2020language}, we show that  ``I am going to submit this paper." is twice more likely to occur compared to ``I I am going to submit this paper." We used the different GPT models (Neo variants and GPT3) \cite{brown2020language} as our oracle. All models yield directionally similar results. However, the larger models lead to the lower and more discriminative relative scoring. We thus refer to our metric as TRScore-GPT in the context of our results. The reference code for TRScore would be shared widely upon institutional approval. Alternative models include BLOOM\cite{bigscience_workshop_2022}, OPT\cite{zhang2022opt}, and T-NLG\cite{smith2022using}.

We first take the reference transcription and compute the likelihood scores of all sentences. In the absence of reference transcriptions, we can use a high-quality news corpora or any alternative corpus that one may want to compare against (especially in the context of A/B testing).
In our experience, to get a robust estimate one should try to have $\sim$100-150 sentences to compare against.
The 50\textsuperscript{th} percentile likelihood score of this distribution is considered the baseline score.
We generate a similar percentile distribution for speech transcripts from our candidate systems. TRScore is then computed relative to the baseline score:

\mathchardef\mhyphen="2D
\[TRScore_{x} = \frac{{P\mhyphen 50}_{reference}}{{P\mhyphen x}_{candidate}} \times 100\%\]

It is worth noting that even though the lower bound of TRScore is 0\%, TRScore could go higher than 100\%, as would be the case typically for TRScore$_{25}$. This is by design, and indicates that some sentences may be more readable than the median reference sentence. Similarly, TRScore$_{90}$\textsuperscript{ref} would be much lower than 100\%, indicating that w.r.t our reference text, some sentences are less readable than the median.

\subsection{Human Readability Score}
We asked six language experts to score sentences from each output variants on a 5-point scale as shown in Table \ref{tab:human_readability_scale}. We then normalized and aggregated the scores into one human readability score (HRS) ($0\%$ to $100\%$) per set.

\begin{table}[!hb]
  \centering
  \begin{tabular}{|l|l|}
    \hline
    \textbf{Rating} & \textbf{Description}\\
    \hline
    0 & Incomprehensible \\
    1 & Grammatically incorrect ($> 1$ error) \\& but only partially comprehensible\\
    2 & Grammatically incorrect ($> 1$ error) \\& but still comprehensible \\
    3 & Grammatically somewhat correct ($> 0$ errors) \\& but comprehensible \\
    4 & Grammatically correct and comprehensible \\
    \hline
  \end{tabular}
  \caption{Judge guidelines for readability scoring}
  \label{tab:human_readability_scale}
\end{table}

\subsection{Experiments}
\label{sec:experiment}

To test the versatility and reliability of TRScore, we evaluated candidate systems across various long-form transcription scenarios. All our test sets have audio and associated written form transcription. We calculated punctuation F$_1$-scores against the reference transcriptions, and compared them with TRScore-GPT for all setups. For a small subset of systems, we also collected human readability scores and compared them with TRScore$_{50}$-GPT.


Many transcription features impact readability. Examples include disfluency handling, capitalization, and inverse text normalization. For our experiments, we focus on analyzing the effectiveness of TRScore at capturing the impact of punctuation alone on readability. Our candidate systems vary in their ASR decoder segmentation models and punctuation models. We keep the rest of the system constant, including the acoustic model (AM) and the language model (LM). This ensures that the candidate systems have similar word error rates (WER).

\newcolumntype{Y}{>{\centering\arraybackslash}X}
\begin{table*}[!t]
  \centering
  \begin{tabularx}{\textwidth}{|c|c|Y|Y|Y|Y|Y|Y|} 
    \hline
    \multirow{2}{*}{\textbf{Test Set}} &
    \multirow{2}{*}{\centering \textbf{Model}} &
    \multicolumn{3}{|c|}{\textbf{Punctuation}} &
    \multicolumn{3}{|c|}{\textbf{TRScore-GPT-3 \textit{(w.r.t. P$_{50}$-reference)}}}\\
    \cline{3-8}
    & & P & R & F$_{1}$ & P$_{50}$ & P$_{75}$ & P$_{90}$\\
    \hline
    \multirow{2}{*}{\textit{NPR-76}}
    & {Baseline} & 72 & 69 & 71 & 69 & 34 & 14\\
    & {Streaming} & 79 & 71 & 75 & 81 & 40 & 18\\
    \hline
    \multirow{2}{*}{\textit{EP-100}}
    & {Baseline} & 57 & 61 & 59 & 60 & 29 & 10\\
    & {Streaming} & 64 & 60 & 62 & 72 & 37 & 18\\
    \hline
    \multirow{2}{*}{\textit{Earnings-10}}
    & {Baseline} & 67 & 63 & 65 & 62 & 31 & 14\\
    & {Streaming} & 72 & 62 & 66 & 71 & 37 & 19\\
    \hline
    \multirow{2}{*}{\textit{Dictation-100}}
    & {Baseline} & 68 & 63 & 65 & 62 & 29 & 11\\
    & {Streaming} & 74 & 60 & 67 & 76 & 37 & 17\\
    \hline
    \multirow{2}{*}{\textit{Voicemail-400}}
    & {Baseline} & 67 & 60 & 63 & 82 & 41 & 19\\
    & {Streaming} & 72 & 54 & 62 & 97 & 49 & 23\\
    \hline
  \end{tabularx}
  \caption{TRScore-GPT-3 can be used to select punctuation models/systems in the absence of reference transcription. A score of 100 is attributed to reference text TRScore$_{50}$-GPT-3.}
  \label{tab:punctuation}
\end{table*}

\subsubsection{ASR Decoder Segmentation Models}
We consider 3 candidate segmentation systems explored previously \cite{behre2022smart}:\newline

\noindent
\textbf{VAD-only}: Voice Activity Detection in silence based segmentation with a 500ms threshold

\noindent
\textbf{RNN-EOS}: Recurrent Neural Network (RNN) based Language Model end-of-segment predictor (LM-EOS) and Voice Activity Detection end-of-segment predictor (VAD-EOS) models with no look-ahead

\noindent
\textbf{RNN-EOS look-ahead}: RNN based LM-EOS and VAD-EOS models with one-word look-ahead

\subsubsection{Punctuation Models}
We consider two candidate punctuation systems \cite{behre2022streamingpunc}:\newline

\noindent
\textbf{Baseline}: Transformer punctuation model, always punctuating at segment boundaries.

\noindent
\textbf{Streaming}: Streaming punctuation with transformers, which leverages bidirectional context across segment boundaries.

\subsubsection{Test sets}
We evaluate our model performance across various scenarios using both public and internal test sets.
\newline\newline
\noindent
\textbf{NPR-76} \cite{npr}: 20 hours of test data from 76 transcribed NPR Podcast episodes.

\noindent
\textbf{EP-100} \cite{europar}: This dataset contains 100 English sessions scraped from European Parliament Plenary videos.

\noindent
\textbf{Earnings-10}: 10 hours of earnings call transcription data from the MAEC corpus \cite{li2020maec}

\noindent
\textbf{Dictation-100}: This internal set consists of 100 utterances with human transcriptions.

\noindent\textbf{Voicemail-400}: This internal set consists of 400 voicemails with human transcriptions.

\section{Results}
\label{sec:results}

In Table \ref{tab:human}, we present human readability scores with standard deviations and the corresponding TRScore$_{50}$-GPT scores across several ASR decoder segmentation systems. TRScore$_{50}$-GPT is strongly correlated with human readability score, with a Pearson's \textit{r} correlation coefficient of 0.98. Here, we keep the acoustic, language, and punctuation models fixed, so lexical WER is roughly the same across systems. Thus, the observed differences in readability are solely attributed to variation in ASR decoder segmentation techniques.

\begin{table}[!ht]
  \centering
  \begin{tabular}{|c|c|c|}
    \hline
    \textbf{Model} & \textbf{HRS} & \textbf{TRScore$_{50}$}\\
    \hline
    VAD-only & \begin{math}75\% \pm 19.1\%\end{math} & 59\%\\
    \hline
    RNN-EOS & \begin{math}74\% \pm 23.3\%\end{math} & 61\%\\
    \hline
    RNN-EOS look-ahead & \begin{math}79\% \pm 21.0\%\end{math} & 69\%\\
    \hline
    Human transcriptions & \begin{math}86\% \pm 10.4\%\end{math} & 100\%\\
    \hline
  \end{tabular}
  \caption{TRScore$_{50}$-GPT vs Human Readability Scores across ASR decoder segmentation systems}
  \label{tab:human}
\end{table}

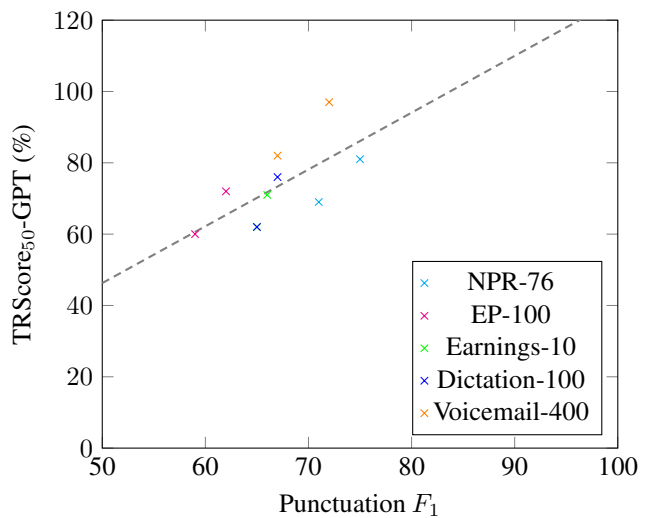
\begin{figure}[!ht]
\begin{tikzpicture}
      \begin{axis}[
        xlabel={Punctuation $F_1$},
        ylabel={TRScore$_{50}$-GPT (\%)},
        legend pos=south east,
        xmin=50, xmax=100,
        ymin=0, ymax=120,
        xtick={50,60,...,100},
        xticklabels={50,60,...,100},   
        ytick={0,20,...,120},
        domain=50:100
                ]
        \addplot[only marks, color=cyan, mark=x]
        coordinates{ 
          (71,69)     
          (75,81)     
        };
        \addlegendentry{NPR-76}
        
        \addplot[only marks, color=magenta, mark=x]
        coordinates{ 
          (59,60)     
          (62,72)     
        };
        \addlegendentry{EP-100}

        \addplot[only marks, color=green, mark=x]
        coordinates{ 
          (65,62)     
          (66,71)     
        };
        \addlegendentry{Earnings-10}
        
        \addplot[only marks, color=blue, mark=x]
        coordinates{ 
          (65,62)     
          (67,76)     
        };
        \addlegendentry{Dictation-100}
        
        \addplot[only marks, color=orange, mark=x]
        coordinates{ 
          (67,82)     
          (72,97)     
        };
        \addlegendentry{Voicemail-400}
        
        \addplot+[no marks,gray,thick] {1.5929 * x - 33.37 } ;    
      \end{axis}
        
    \end{tikzpicture}
    \caption{Correlation: Punctuation $F_1$ and TRScore$_{50}$-GPT}
    \label{fig:correlationwithpunc}
\end{figure}

Table \ref{tab:punctuation} shows punctuation F$_1$, computed based on written-form reference text and aggregated across periods, commas and question marks. TRScore$_{50}$-GPT is strongly correlated with punctuation $F_1$, with a Pearson's \textit{r} correlation coefficient of 0.67 as observed in Figure \ref{fig:correlationwithpunc}.

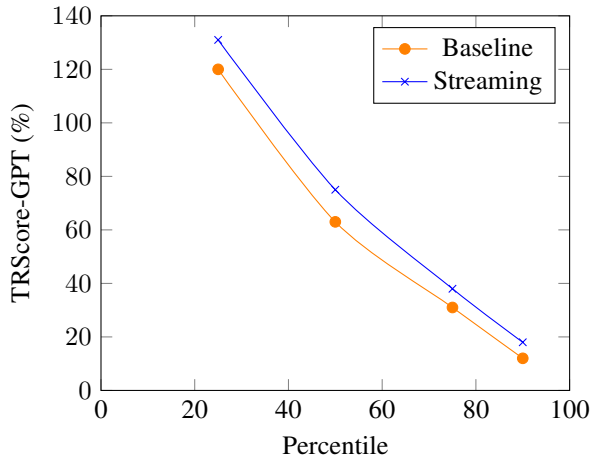
\begin{figure}[!ht]
    \begin{tikzpicture}
    \pgfplotsset{%
        width=0.35\textwidth,
        height=0.28\textwidth,
          scale only axis,
      }
\begin{axis}[
    legend pos=north east,
    ylabel={TRScore-GPT (\%)},
    xlabel={Percentile},
    xmin=0, xmax=100,
    ymin=0, ymax=140,
    xtick={0,20,...,100},
    xticklabels={0,20,...,100}, 
    ytick={0,20,...,140},
    domain=0:100
            ]
\addplot[smooth,mark=*,orange] plot coordinates {
    (25, 120)
    (50, 63)
    (75, 31)
    (90, 12)
};
\addlegendentry{Baseline}

\addplot[smooth,color=blue,mark=x]
    plot coordinates {
    (25, 131)
    (50, 75)
    (75, 38)
    (90, 18)
    };
\addlegendentry{Streaming}

\end{axis}
    \end{tikzpicture}
    \caption{Comparing readability across percentiles}
    \label{fig:percentilechart}
\end{figure}

Figure \ref{fig:percentilechart} shows the TRScore-GPT percentile distribution for two candidate punctuation systems aggregated across datasets. This showcases the reliability and stability of the TRScore metric across percentiles.

\section{Discussion}
\label{sec:discussion}

Our comparison of human readability scoring and TRScore$_{50}$-GPT shows the high standard deviation in human scores (high inter-observer variability), a problem that makes it difficult to ascertain whether differences in averaged human ratings between systems are meaningful or not. By comparison, TRScore$_{50}$-GPT more definitively measures the readability of each system and helps determine whether VAD-only or RNN-EOS offers more readable segmentation results. In this way, TRScore is helpful not only as a standalone proxy for human judgment, but also as a deterministic second opinion when human judgment is available.

In addition to human scores, we establish a clear correlation between TRScore-GPT and punctuation F$_1$ scores.
In the absence of human judgment between systems, punctuation F$_1$ is the go-to evaluation metric used to measure models' relative performance.
However, to calculate punctuation F$_1$ on any given set, we need fully punctuated written-form human transcriptions as our ground truth labels. Producing such transcriptions requires manual human labor, and the quality of the human transcriptions may vary, just as human judgments do in human scoring. For conversational speech with lots of disfluencies, the quality of human punctuation varies wildly.
With TRScore, we no longer require human transcriptions.
Because TRScore-GPT is strongly correlated with both human judgment and punctuation F$_1$, TRScore is a powerful way to compare models in the absence of human transcriptions and human scorers.
Furthermore, the standalone nature of TRScore makes it especially useful in compliant environments where scientists cannot log or analyze the speech recognition outputs.

\begin{table}[!ht]
  \centering
  \begin{tabular}{|l|c|}
    \hline
    \centering
    \textbf{Reference / Hypothesis Text} & \textbf{TRScore-GPT}\\
    \hline
    I am going to submit this paper. & 100\%\\
    I am going to\textbf{. Submit} this paper. & 32\%\\
    \hline
    I am going to submit this paper. & 100\%\\
    I am going\textbf{. To} submit this paper. & 9\%\\
    \hline
    I am going to submit this paper. & 100\%\\
    I am going to \textbf{umm} submit this paper. & 30\%\\
    \hline
    I am going to submit this paper to ICASSP. & 100\%\\
    I am going to submit this paper to \textbf{icassp}. & 57\%\\
    \hline
    Submit this paper to ICASSP at 7 AM. & 100\%\\
    Submit this paper to ICASSP at \textbf{seven am}. & 56\%\\
    \hline
  \end{tabular}
  \caption{Qualitative examples of text and corresponding TRScore-GPT-3 (with respect to each reference text). TRScore-GPT-3 captures important aspects of readability such as punctuation, disfluency, capitalization, and ITN.}
  \label{tab:puncexamples}
\end{table}

Thus far, we have primarily focused our attention on the impact of punctuation and segmentation on readability, as measured by TRScore-GPT.
However, as we present in Table \ref{tab:puncexamples}, TRScore-GPT also captures the readability impact of other elements such as disfluency, capitalization, and ITN formatting.
A key limitation of metrics such as WER and SER is that they weigh each word equally rather than measuring overall readability.
TRScore-GPT, however, is able to distinguish between two identical texts with just one display error differentiating them.
The first example in Table \ref{tab:puncexamples} may indeed contain a punctuation error, but the two sentences individually are still grammatically valid.
This is not the case, however, in the second example.
TRScore-GPT captures this distinction, scoring the second example as less readable than the first.
In the other three examples, our metric accurately detects that text is less readable when disfluencies or errors in capitalization or ITN are present.

\section{Conclusion}
\label{sec:format}

Segmentation and punctuation have become increasingly important for ASR systems. The traditional metrics of precision, recall, and F$_1$ require high quality written-form transcriptions. However, acquiring such transcriptions is a slow and expensive endeavor and the results often suffer from significant inter-judge variability. In this paper, we introduced a novel automatic readability metric which leverages large-scale pre-trained models like GPT-3 to score ASR outputs. We demonstrated TRScore-GPT's strong correlation with human readability scores as well as punctuation F$_1$ scores, over a variety of ASR use cases. TRScore is not only makes a more powerful, cheaper, and faster alternative to human evaluation, but also a viable option for compliance or enterprise use cases in which logging data or acquiring human transcription is highly restricted. We make our tool to the speech community for readability scoring purposes.

\FloatBarrier

\vfill\pagebreak

\bibliographystyle{IEEEbib}
\bibliography{refs}

\begin{thebibliography}{10}

\bibitem{sellam2020bleurt}
Thibault Sellam, Dipanjan Das, and Ankur Parikh,
\newblock ``{BLEURT}: Learning robust metrics for text generation,''
\newblock in {\em Proceedings of the 58th Annual Meeting of the Association for
  Computational Linguistics}, Online, July 2020, pp. 7881--7892, Association
  for Computational Linguistics.

\bibitem{rei2020comet}
Ricardo Rei, Craig Stewart, Ana~C Farinha, and Alon Lavie,
\newblock ``{COMET}: A neural framework for {MT} evaluation,''
\newblock in {\em Proceedings of the 2020 Conference on Empirical Methods in
  Natural Language Processing (EMNLP)}, Online, Nov. 2020, pp. 2685--2702,
  Association for Computational Linguistics.

\bibitem{stanojevic2014beer}
Milo{\v{s}} Stanojevi{\'c} and Khalil Sima’an,
\newblock ``Beer: Better evaluation as ranking,''
\newblock in {\em Proceedings of the Ninth Workshop on Statistical Machine
  Translation}, 2014, pp. 414--419.

\bibitem{papineni2002bleu}
Kishore Papineni, Salim Roukos, Todd Ward, and Wei-Jing Zhu,
\newblock ``Bleu: a method for automatic evaluation of machine translation,''
\newblock in {\em Proceedings of the 40th annual meeting of the Association for
  Computational Linguistics}, 2002, pp. 311--318.

\bibitem{popovic-2015-chrf}
Maja Popovi{\'c},
\newblock ``chr{F}: character n-gram {F}-score for automatic {MT} evaluation,''
\newblock in {\em Proceedings of the Tenth Workshop on Statistical Machine
  Translation}, Lisbon, Portugal, Sept. 2015, pp. 392--395, Association for
  Computational Linguistics.

\bibitem{rouge}
Chin-Yew Lin,
\newblock ``Rouge: A package for automatic evaluation of summaries,''
\newblock in {\em Association for Computational Linguistics}, 2004, pp. 74--81.

\bibitem{zhao2019moverscore}
Wei Zhao, Maxime Peyrard, Fei Liu, Yang Gao, Christian~M. Meyer, and Steffen
  Eger,
\newblock ``{M}over{S}core: Text generation evaluating with contextualized
  embeddings and earth mover distance,''
\newblock in {\em Proceedings of the 2019 Conference on Empirical Methods in
  Natural Language Processing and the 9th International Joint Conference on
  Natural Language Processing (EMNLP-IJCNLP)}, Hong Kong, China, Nov. 2019, pp.
  563--578, Association for Computational Linguistics.

\bibitem{zhang2019bertscore}
Tianyi Zhang, Varsha Kishore, Felix Wu, Kilian~Q Weinberger, and Yoav Artzi,
\newblock ``Bertscore: Evaluating text generation with bert,''
\newblock {\em arXiv preprint arXiv:1904.09675}, 2019.

\bibitem{yuan2021bartscore}
Weizhe Yuan, Graham Neubig, and Pengfei Liu,
\newblock ``Bartscore: Evaluating generated text as text generation,''
\newblock {\em Advances in Neural Information Processing Systems}, vol. 34, pp.
  27263--27277, 2021.

\bibitem{kim2021evaluating}
Suyoun Kim, Duc Le, Weiyi Zheng, Tarun Singh, Abhinav Arora, Xiaoyu Zhai,
  Christian Fuegen, Ozlem Kalinli, and Michael~L. Seltzer,
\newblock ``Evaluating user perception of speech recognition system quality
  with semantic distance metric,''
\newblock in {\em INTERSPEECH}, 2022.

\bibitem{cognitiveopenai}
``Cognitive services: Open ai service,''
  \url{https://azure.microsoft.com/en-us/products/cognitive-services/openai-service/#features},
\newblock Accessed: 2022-10-23.

\bibitem{brown2020language}
Tom Brown, Benjamin Mann, Nick Ryder, Melanie Subbiah, Jared~D Kaplan, Prafulla
  Dhariwal, Arvind Neelakantan, Pranav Shyam, Girish Sastry, Amanda Askell,
  et~al.,
\newblock ``Language models are few-shot learners,''
\newblock {\em Advances in neural information processing systems}, vol. 33, pp.
  1877--1901, 2020.

\bibitem{bigscience_workshop_2022}
{BigScience Workshop},
\newblock ``Bloom (revision 4ab0472),'' 2022.

\bibitem{zhang2022opt}
Susan Zhang, Stephen Roller, Naman Goyal, Mikel Artetxe, Moya Chen, Shuohui
  Chen, Christopher Dewan, Mona Diab, Xian Li, Xi~Victoria Lin, et~al.,
\newblock ``Opt: Open pre-trained transformer language models,''
\newblock {\em arXiv preprint arXiv:2205.01068}, 2022.

\bibitem{smith2022using}
Shaden Smith, Mostofa Patwary, Brandon Norick, Patrick LeGresley, Samyam
  Rajbhandari, Jared Casper, Zhun Liu, et~al.,
\newblock ``Using deepspeed and megatron to train megatron-turing nlg 530b, a
  large-scale generative language model,''
\newblock {\em arXiv preprint arXiv:2201.11990}, 2022.

\bibitem{behre2022smart}
Piyush Behre, Naveen Parihar, Sharman Tan, Amy Shah, Eva Sharma, Geoffrey Liu,
  Shuangyu Chang, Hosam Khalil, Chris Basoglu, and Sayan Pathak,
\newblock ``Smart speech segmentation using acousto-linguistic features with
  look-ahead,''
\newblock {\em arXiv preprint arXiv:2210.14446}, 2022.

\bibitem{behre2022streamingpunc}
Piyush Behre, Sharman Tan, Padma Varadharajan, and Shuangyu Chang,
\newblock ``Streaming punctuation for long-form dictation with transformers,''
\newblock {\em arXiv preprint arXiv:2210.05756}, 2022.

\bibitem{npr}
``Home page top stories,'' \url{https://www.npr.org/},
\newblock Accessed: 2022-05-30.

\bibitem{europar}
``Debates and videos: Plenary: European parliament,''
  \url{https://www.europarl.europa.eu/plenary/en/debates-video.html},
\newblock Accessed: 2022-05-30.

\bibitem{li2020maec}
Jiazheng Li, Linyi Yang, Barry Smyth, and Ruihai Dong,
\newblock ``Maec: A multimodal aligned earnings conference call dataset for
  financial risk prediction,''
\newblock in {\em Proceedings of the 29th ACM International Conference on
  Information \& Knowledge Management}, 2020, pp. 3063--3070.

\end{thebibliography}

\end{document}